\documentclass{article}

\PassOptionsToPackage{numbers, compress}{natbib}
\usepackage[final]{nips_2016}

\usepackage[utf8]{inputenc} 
\usepackage[T1]{fontenc}    
\usepackage{hyperref}       
\usepackage{url}            
\usepackage{booktabs}       
\usepackage{amsfonts}       
\usepackage{nicefrac}       
\usepackage{microtype}      

\usepackage{color}
\usepackage[pdftex]{graphicx}
\graphicspath{{pics/}{../pdf/}{../jpeg/}}
\DeclareGraphicsExtensions{.pdf,.jpeg,.png}

\title{Towards end-to-end optimisation \\ of functional image analysis pipelines}

\author{
 Albert Vilamala\\
 Technical University of Denmark\\
 \texttt{alvmu@dtu.dk} \\
  \And
  Kristoffer Hougaard Madsen \\
  Danish Research Centre for Magnetic Resonance\\
  \texttt{kristofferm@drcmr.dk} \\
  \And
   Lars Kai Hansen \\
   Technical University of Denmark\\
   \texttt{lkai@dtu.dk}
}

\begin{document}

\maketitle

\begin{abstract}
The study of neurocognitive tasks requiring accurate localisation of activity often rely on functional Magnetic Resonance Imaging, a widely adopted technique that makes use of a pipeline of data processing modules, each involving a variety of parameters. These parameters are frequently set according to the local goal of each specific module, not accounting for the rest of the pipeline. Given recent success of neural network research in many different domains, we propose to convert the whole data pipeline into a deep neural network, where the parameters involved are jointly optimised by the network to best serve a common global goal. As a proof of concept, we develop a module able to adaptively apply the most suitable spatial smoothing to every brain volume for each specific neuroimaging task, and we validate its results in a standard brain decoding experiment.
\end{abstract}

\section{Introduction}
\begin{figure}[htb!]
\centering
\includegraphics[width=0.6\textwidth]{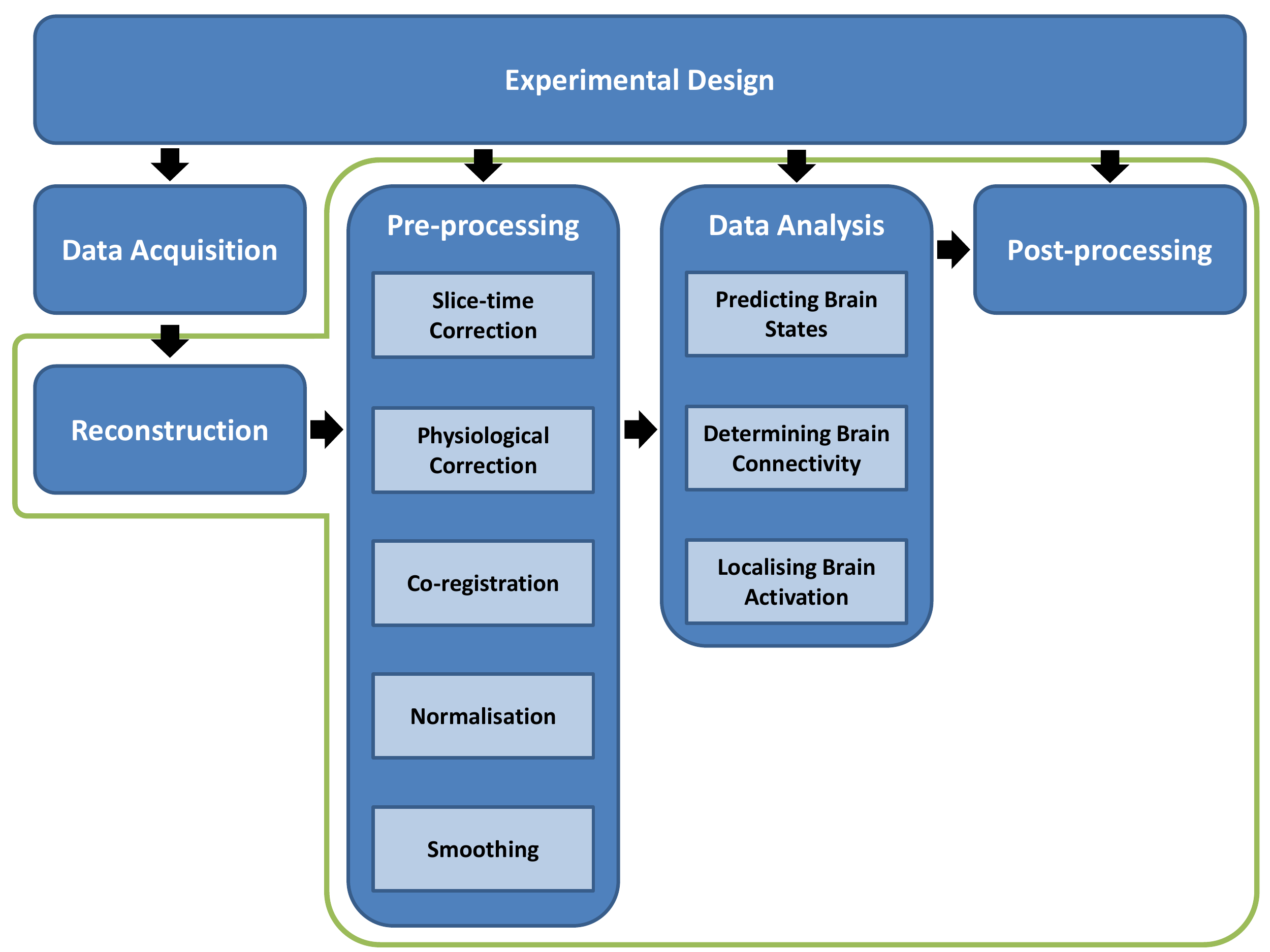}
\caption{Data processing pipeline for fMRI.}
\label{fig:fmri_pipeline}
\end{figure}

Functional Magnetic Resonance Imaging (fMRI) has been the leading neuroimaging technique for monitoring brain activity in a non-invasive manner, when high spatial resolution is required. Nevertheless, direct usage of raw data from the scanner is not appropriate as a measure of neuronal activity and a comprehensive data processing pipeline is often applied \cite{Lindquist2008}. As depicted in Fig.~\ref{fig:fmri_pipeline}, a standard fMRI pipeline might consist of six differentiated modules: in the \emph{experimental design}, the whole experimental framework is accurately planned for the specific concept of investigation; then subjects are invited to perform specific tasks while lying in a scanner, capturing the brain signals during the \emph{data acquisition} phase; afterwards, in the \emph{reconstruction} step, data retrieved in the frequency domain are converted to time domain; next, data undergo \emph{preprocessing}, where various modules with different purposes (e.g. eliminating artifacts, acquisition errors or brain normalisation) are chained together; finally \emph{data analysis} can be carried out (e.g. localising brain activity, connectivity or decoding); in some cases additional \emph{post processing} can be performed in order to refine the obtained results. 

This approach has proven useful in many settings, and now serve as a gold standard for fMRI data processing, leading to the development of a wide range of different methodologies and tools responsible for each step of the aforementioned pipeline \cite{Friston1995, Smith2004, Cox1996}  However, there is still a major design caveat making the whole process both challenging and suboptimal: each one of the modules described in Fig.~\ref{fig:fmri_pipeline} performs an isolated task which often requires one or more parameters to be set (usually by hand); hence, requiring human expertise or otherwise use a default value. Notice that these parameters do not only have great impact to their specific module but also to subsequent processing components \cite{Churchill2012, Churchill2012b}.

Deep neural networks represent the current state of the art learning algorithms; their popularity being devoted to the last achievements in a variety of fields as object recognition \cite{Krizhevsky2012, Simonyan2014, Szegedy2015} or natural language processing \cite{Graves2013, Sutskever2014}, among others. All these improvements are made possible by end-to-end optimisation of the pipelines... Our question is: can this be transferred to the neuroimaging domain?
Roughly speaking, a neural network can be seen as a deep structure made up of different blocks arranged in layers, which are assembled to perform a specific common task. The components in such a heterogeneous architecture can be of varying nature according to the subtask they are addressing; however, they usually share a common ability to adapt to the current problem by means of a set of weights that are iteratively being optimised. The optimisation process aims at finding a configuration of weights that best deals with the global task and not each subtask independently; hence, there exists a flow of information through the network accounting for the performance of each component with respect to the global goal.

By confronting this definition with the fMRI pipeline structure, we hypothesise that results obtained using the fMRI pipeline can be improved if the whole pipeline (i.e. all steps in the green box in Fig.~\ref{fig:fmri_pipeline}) is transformed into a deep heterogeneous neural network, allowing the parameters of each component to be globally optimised for a given task.

Accordingly, we here present our first contribution to generically turn the fMRI pipeline into a deep heterogeneous network by designing a specific feed forward network architecture. As a proof of concept, a module aiming at replacing the crucial smoothing step in the fMRI pipeline is provided; a component that not only offers easy training using standard backpropagation, but it also adaptively chooses the most adequate degree of smoothing to apply to each brain volume for a specific neuroimaging task.

\section{Model}
Let $S = \left\{\mathbf{X}_n\right\}_{n=1}^N$ be a set containing fMRI brain volumes, each defined as $\mathbf{X} \in \mathbb{R}^{H \times W \times D}$ of height $H$, width $W$ and depth $D$. Our goal is to learn a function $f(\mathbf{X})$ encoding all the operations to be applied to $\mathbf{X}$ in order to conduct a specific neuroimaging task; that is, $f$ might define the whole data processing pipeline in Fig.~\ref{fig:fmri_pipeline} or part of it.

\begin{figure}[htb!]
\centering
\includegraphics[width=0.9\textwidth]{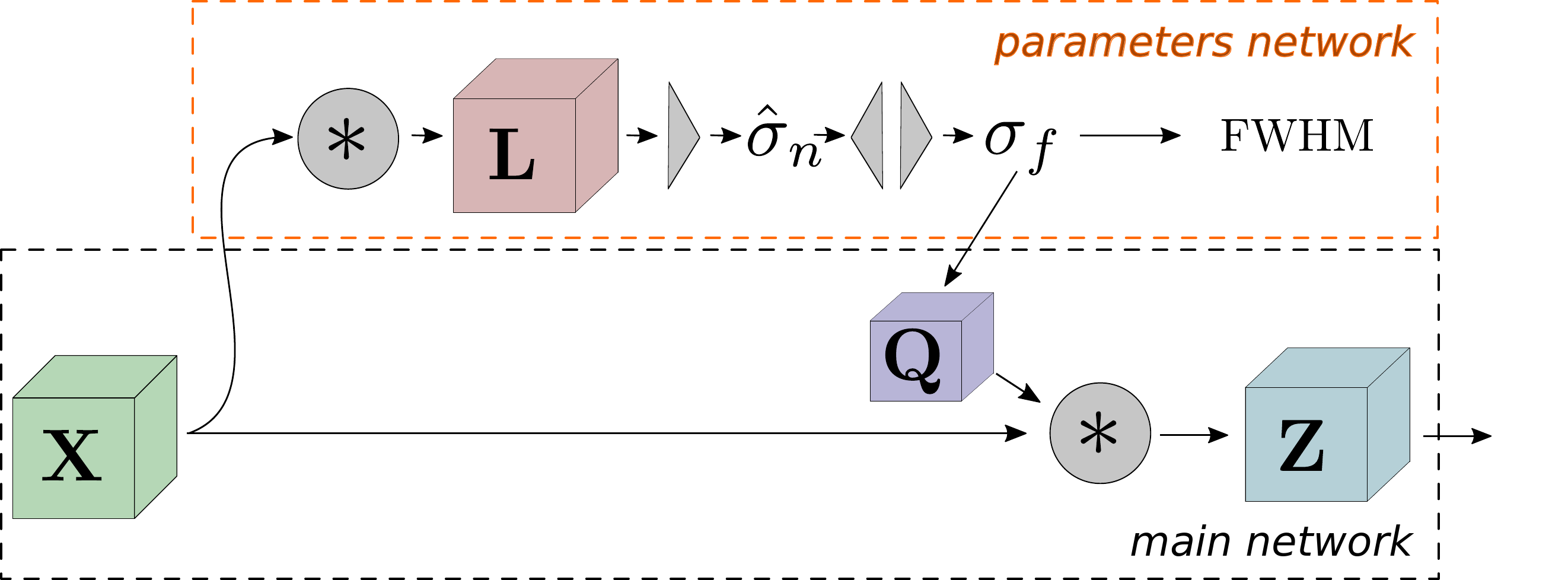}
\caption{Neural network architecture for an fMRI data processing pipeline.}
\label{fig:architecture}
\end{figure}

Inspired by the work in \cite{Jaderberg2015}, our proposal consists in learning $f$ using a neural network made up of two subnetworks (Fig.~\ref{fig:architecture}): the \emph{main network}, which performs the actual transformation of the input brain volume; and an external network, called the \emph{parameters network}, which calculates the most adequate values for the parameters required by any of the components in the main network. Such architecture allows joint optimisation of existing parameters in the two subnetworks by end-to-end learning.

In the particular case of building an adaptive smoothing component, our goal is to learn $f$ that optimises the degree of smoothing to be applied to each brain volume for a particular neuroimaging task. In this respect, the \emph{main network} will produce a smoothed version of the input volume by spatially convolving it with a Gaussian filter, the width of which will be calculated by the \emph{parameters network}.

\subsection{The main network}
The task of this network is to perform all the required operations to the input data assisted by the parameters generated by the \emph{parameters network}. In this case, given an input volume and a value specifying the most suitable standard deviation parameter of the Gaussian filter ($\sigma_f$), the \emph{main network} first constructs the appropriate filter and it convolves it with the input. All these steps are detailed next.

\subsubsection{Constructing the Gaussian filter}
Given a value for the $\sigma_f$ parameter, we define the appropriate smoothing function parameterised as a continuous isotropic 3-dimensional Gaussian function:
\begin{equation}
g(x,y,z; \sigma_f) = \frac{1}{\left(\sqrt{2\pi} \sigma_f\right)^3} \exp\left\{-\frac{x^2 + y^2 + z^2}{2\sigma_f^2}\right\}
\label{eq:gaussian}
\end{equation}

In order to apply the smoothing we here consider a discrete approximation of the filter in a sampling grid, whose size is determined by $\sigma_f$. In other words, let $G$ be a gird, such that $G_i=(x_i, y_i, z_i)$, where $-\lfloor(t \cdot \sigma_f + 0.5)/2\rfloor \leq x_i, y_i,z_i \leq \lfloor(t \cdot \sigma_f + 0.5)/2\rfloor$; $t$ being a parameter specifying the number of standard deviations from the mean where the Gaussian function will be truncated.


The final $\mathbf{Q}$ filter is obtained by sampling from Eq.~\ref{eq:gaussian} at the specific locations specified by $G$ and renormalising its values, such that the sum of them all adds to $1$.

Several important implementation details worth to mention are concerned with the use of low $\sigma_f$ values: whenever $\sigma_f < 1.5/t$, the truncated discrete Gaussian function generates a single-cell $\mathbf{Q}$ filter, meaning that all partial derivatives of $\mathbf{Q}_{i,j,k}$ are $0$; therefore, causing the backpropagation gradient to disappear. This misbehaviour is alleviated by stochastically adding $1.0$ to $\sigma_f$ with probability $p$ for values below the mentioned threshold. Moreover, special attention must be paid in this regime when renormalising the single-cell $\mathbf{Q}$ filter, since dividing by the sum of all elements cancels $\sigma_f$ out, also causing the gradient to disappear. 

\subsubsection{Volume smoothing}
Given a convolutional smoothing operator $\mathbf{Q} \in \mathbb{R}^{H' \times W' \times D'}$, where $H' < H$, $W' < W$, $D' < D$, the process of smoothing a volume can be accomplished by sliding the operator over the brain volume, performing a convolution in each location. The resulting smoothed volume $\mathbf{Z} \in \mathbb{R}^{H \times W \times D}$ is computed as:
\[
\mathbf{Z}_{h,w,d} = \sum_{i=0}^{H'}\sum_{j=0}^{W'}\sum_{k=0}^{D'} \mathbf{X}_{h+i,w+j,d+k} \cdot \mathbf{Q}_{i,j,k}
\]


At this point $\mathbf{Z}$ can be used to subsequent analysis, for instance brain decoding. However, up to now we have been omitting the origin of the important $\sigma_f$ parameter in the Gaussian filter, which is calculated using the \emph{parameters network}.

\subsection{The parameters network}
The aim of this network is to compute the optimal values for the parameters required by the \emph{main network}. In our case, calculating the most appropriate $\sigma_f$ value for the Gaussian filter, which is obtained by first estimating the average noise in the whole brain volume ($\hat{\sigma}_n$) and converting it to $\sigma_f$ in voxel units.

\subsubsection{Noise estimation}
Based on \cite{Immerkaer1996}, a fast procedure to estimate the noise in a volume can be carried out by convolving the input with a Laplacian filter, averaging the absolute value of the resulting volume and scaling. In our context, this same operation can be obtained by sequentially applying conventional neural network layers, such as convolution (using Laplacian filter) with absolute value non-linearity followed by 3D average pooling of the whole volume, plus fully connected layers for scaling.

\subsubsection{Mapping from $\hat{\sigma}_n$ to $\sigma_f$}
Laplacian convolution and average pooling of absolute values are pretty straightforward. The somewhat more involved part is to convert from noise estimate $\hat{\sigma}_n$ to the actual $\sigma_f$ of the Gaussian filter. To do so, we allow the fully connected layers mentioned in the previous paragraph to account not only for the scaling factor, but also for the $\hat{\sigma}_n$ to $\sigma_f$ mapping. More precisely, a first linear fully connected layer containing one input and $M$ outputs is in charge of producing several linear modifications to $\hat{\sigma}_n$ and a second fully connected layer with an exponential activation function combines those variations into a single positive value $\sigma_f$.


\section{Empirical evaluation}
In this section, we evaluate the proposed \emph{adaptive smoothing} network in a brain decoding task. 
We here consider fMRI data from a simple sequential finger tapping
paradigm, in which subjects did alternated between 20 second blocks of
left and right finger tapping separated by 10 seconds of rest. Data
was recorded on a Siemens 3T scanner (Magnetom Trio) equipped with a
standard birdcage headcoil. Each of the 29 subjects' data consisted of
240 volumes with 3 mm isotropic resolution sampled at a repetition
time of 2.49 seconds. Further acquisition parameters can be found in
\cite{Rasmussen2011}
which analysed the same data. After basic preprocessing steps
including realignment and normalisation by standard settings in SPM12
(\url{http://www.fil.ion.ucl.ac.uk/spm/software/spm12}) we labelled each volume according to the left
right condition rejecting the first volume in each block.
The goal of our network is to correctly classify these two distinct conditions when different levels of white noise have been artificially added to the input data. We expect our network to correct for the noise by applying the most appropriate level of Gaussian smoothing to each brain volume leading to an improvement in accuracy as compared to the baseline case, where a fixed filter is used.


\subsection{Experimental setup}
Data have been normalised within the $[0,1]$ interval by using the overall intra-subject extrema. Then, we have split the $29$ subjects into three categories: $21$ for training, $4$ for validation and $4$ for test. Notice that all volumes of a specific subject fall entirely into the same category. It is important to realise at this point that all scans for the same subject are analysed together, allowing us to apply intra-subject normalisation. This has implications to the number of brain volumes used in our training mini-batch (i.e. $120$) as well as the way we implement batch-normalisation.

Several noisy versions of each brain volume have been artificially generated by adding random zero-mean Gaussian noise of varying intensity, $\sigma = \{ 0.1, 0.2, 0.3\}$, to every voxel. 

The same network as the one depicted in Fig.~\ref{fig:architecture} has been used for this experiment, adding a fully connected layer to the end of it containing a single output node with a sigmoid activation function, acting as a classifier. Weights for this layer have been initialised according to Xavier's initialisation \cite{Glorot2010}. Importantly, similar to the successful batch-normalisation technique \cite{Ioffe2015}, data outputted by the last fully connected layer is being standardised by removing the mean and dividing by the standard deviation of the whole mini-batch, right before applying the sigmoid activation. The difference with respect to the original batch-normalisation is that, given the nature of the experiments, we can make use of the current batch statistics for training, as well as validation and testing. Such normalisation is highly recommended, since the neuron easily saturates given the $[0,1]$ input data normalisation. Other important design decisions involve the $\hat{\sigma}_n$ to $\sigma_f$ mapping, where $M=50$ and weights' initialisations have been sampled from $\mathcal{N}(0,0.09)$.

The network has been trained by optimising the binary cross-entropy between predictions and targets using mini-batch stochastic gradient descent, where each mini-batch comprises $120$ brain volumes of the same subject at same noise level. Regularization has been applied by using validation-based early stopping as well as L2-norm on the weights of the final layer. Values for the learning rate and regularisation parameter have been set according to grid search on a logarithmic scale.

\subsection{Results}

\begin{table}[t]
  \caption{Classification accuracy on test set for different noise levels according to the specified FWHM. For the adaptive smoothing column, noise-level FWHM average is specified within parentheses.}
  \label{tab:results}
  \centering
  \begin{tabular}{cllll}
    \toprule
    &\multicolumn{4}{c}{FWHM}                   \\
    \cmidrule{2-5}
    Noise & 3.0 & 8.0 & 13.0 & Adaptive \\
    \midrule
    0.0 & 96.1  & 98.3 & 98.1 & 93.9 (7.4)    \\
    0.1 & 58.6 & 72.8  & 74.7 & 72.8 (8.4)     \\
    0.2 & 51.1  & 55.3 & 63.9 & 63.9 (9.7)   \\
    0.3 & 55.3  & 52.5 & 58.9 & 59.7 (11.2) \\
    \bottomrule
  \end{tabular}
\end{table}

Table~\ref{tab:results} shows the classification accuracy obtained at every noise level on the test set for a network trained on all noise levels and fixed Gaussian smoothing (i.e. FWHM set to $3.0$, $8.0$ or $13.0$ mm; see next section for details on this parameter); last column contains the accuracy achieved by our proposed network, which applies a specific smoothing for each volume (the noise-specific average value shown between parentheses).

As expected, a higher noise level requires a wider smoothing in order to maintain classification performance: this can be observed in the first column (FWHM=$3.0$), where the classification accuracy drops rapidly as soon as a bit of noise is added to the signal. Likewise, the more noise is added to the signal, the wider the required smoothing (second and third columns). Notice that over-smoothing does not seem to affect classification accuracy for the current data. This might be due to the fact that the motor cortex area of the brain involved in left and right movement are relatively far apart, and the applied smoothing does not mix the two signals.

Finally, our proposed network seems to be able to identify the appropriate smoothing for every noise level, obtaining classification accuracy at similar range as the best obtained by fixed smoothing, but avoids excessive smoothing, resulting in sharper more interpretable brain volumes.

\section{Discussion}
Deep neural networks are known to be very flexible algorithms providing high modelling capabilities, which translates to requiring large sets of weights to be optimised. Lots of weights implies big amounts of data to drive the learning, as it is often the case in all success stories in the area \cite{Krizhevsky2012, Simonyan2014, Szegedy2015,Graves2013, Sutskever2014}. In the neuroimaging field, despite recent advancements in creating large standardised freely available repositories \cite{VanHorn2013, Poldrack2013}, access to training data still becomes a major shortcoming. This is a real challenge when desiging the current network, having to use a conventional pipeline as pre-training or applying a priori information when possible; for instance, when constraining the filter to be fully Laplacian (no learning involved) in the \emph{parameters network} or designing the smoothing filter as being of Gaussian shape, instead of freely allowing the filter to be fully learnt from input instances in the \emph{main network}.

Linked to that, the 3D nature of fMRI brain volumes also limits the design of the network, since naive decisions might induce the number of parameters to grow much faster than they would do in other lower dimensional domains. In this respect, structures like fully connected networks on the inputs become prohibitive. 

Another desirable key aspect in our design involves the concept of interpretability of both the resulting output of the network and each intermediate component. This is the rationale behind normalising the input values to the $[0,1]$ interval: they can always be plotted as human-interpretable voxel intensities. Other normalisation procedures can be applied to feed specific components requiring them (e.g. zero-mean normalisation before fully connected layer for classification). 

The \emph{parameters network} plays an important role regarding interpretability, since we would like to compare the values optimised by our network with the ones being used in real practise in order to find similarities and differences that help us to better understand both processes. In the current adaptive smoothing example, the width of the applied Gaussian smoothing is also expressed as \emph{full width at half maximum} (FWHM) in millimetres. 

\section{Conclusions}
In the current paper we showed a general approach to transform a standard fMRI data processing pipeline made up of individually-optimised modules into a deep neural network, where all parameters are jointly optimised to accomplish a common goal. In particular, we have proposed a specific architecture containing a \emph{main network} that performs the actual data transformation, guided by the \emph{parameters network}: a subnetwork in charge of calculating the best values for the parameters required by the \emph{main network}. We have demonstrated the validity of this approach by constructing a module able to adaptively smooth the input volumes in a brain decoding task.

\subsubsection*{Acknowledgments}

This project has received funding from the European Union’s Horizon 2020 research and innovation programme under the Marie Sklodowska-Curie grant agreement No 659860.

\medskip

\bibliographystyle{plain}
\bibliography{references}
%
%

\end{document}